\documentclass[runningheads]{llncs}

\usepackage{eccv}

\usepackage{eccvabbrv}

\usepackage{graphicx}
\usepackage{booktabs}
\usepackage{graphicx}
\usepackage{booktabs}
\usepackage{epsfig}
\usepackage{graphicx}
\usepackage{amsmath}
\usepackage{amssymb}
\usepackage{dsfont}
\usepackage{color}

\usepackage{comment}
\usepackage{multirow, array, booktabs,xspace,bm}
\usepackage{pifont}
\usepackage{amsfonts}
\usepackage{makecell}
\usepackage{dcolumn}
\usepackage{eqparbox}
\usepackage{enumitem}

\newcommand{\cmark}{\ding{51}}%
\newcommand{\xmark}{\ding{55}}%

\usepackage[accsupp]{axessibility}  

\definecolor{cellcol}{gray}{.92}

\usepackage{colortbl}
\definecolor{red2}{rgb}{0.851,0.4,0.4}
\definecolor{blue2}{rgb}{0.266,0.447,0.768}

\usepackage{hyperref}

\usepackage{orcidlink}

\begin{document}

\title{Leveraging Object Priors for Point Tracking} 

\titlerunning{Tracking with Object Priors}

\author{Bikram Boote\inst{1}
\and
Anh Thai\inst{2}
\and
Wenqi Jia\inst{1}
\and
Ozgur Kara \inst{1} 
\and Stefan Stojanov \inst{2} 
\and \\ James M. Rehg \inst{1}\thanks{Corresponding author}
\and Sangmin Lee \inst{1}$^\star$}

\authorrunning{B. Boote et al.}

\institute{University of Illinois Urbana-Champaign, Illinois, USA
\email{\{boote,wenqij5,ozgurk2,jrehg,sangminl\}@illinois.edu}\\
 \and
Georgia Institute of Technology, Georgia, USA\\
\email{\{athai6,sstojanov\}@gatech.edu}
}
\maketitle

\begin{abstract}
 \noindent Point tracking is a fundamental problem in computer vision with numerous applications in AR and robotics. A common failure mode in long-term point tracking occurs when the predicted point leaves the object it belongs to and lands on the background or another object. We identify this as the failure to correctly capture objectness properties in learning to track. To address this limitation of prior work, we propose a novel objectness regularization approach that guides points to be aware of object priors by forcing them to stay inside the the boundaries of object instances. By capturing objectness cues at training time, we avoid the need to compute object masks during testing. In addition, we leverage contextual attention to enhance the feature representation for capturing objectness at the feature level more effectively. As a result, our approach achieves state-of-the-art performance on three point tracking benchmarks, and we further validate the effectiveness of our components via ablation studies. The source code is available at: \href{https://github.com/RehgLab/tracking_objectness}{https://github.com/RehgLab/tracking\_objectness}
  \keywords{Point tracking \and motion analysis \and objectness \and objectness regularization \and contextual attention}
\end{abstract}

\section{Introduction}
\label{sec:intro}

Point tracking, which is the estimation of point correspondences across multiple frames in a video sequence, is a fundamental problem in computer vision. The estimation of point correspondences is fundamental for many tasks in AR/VR \cite{sarlin2022lamar, zhao2020pointar, liu2020arshadowgan}, SfM/SLAM \cite{kerl2013dense, schonberger2016structure, cui2017hsfm}, and autonomous driving \cite{luo2021exploring, li2018stereo, hu2023planning}. {Point tracking also can play a crucial role in instance-level recognition. By accurately tracking points belonging to specific object instances across frames, we can concretely understand instance behavior over time, which can be leveraged in robotics applications involving object manipulation \cite{bharadhwaj2024track2act, heppert2024ditto}}. Point tracking in extended video sequences \cite{pips,pointodyssey,neoral2024mft,tapir,vecerik2023robotap,moing2023dense} is extremely challenging because 1) the appearance of points can change dramatically due to viewpoint, lighting, and shape changes, and 2) points can become occluded and disoccluded over time. Recent particle video-style methods such as PIPS++ \cite{pointodyssey} address these challenges by leveraging multi-frame temporal context windows to improve the robustness of appearance modeling and leverage \emph{temporal continuity} in tracking individual target points. In contrast, optical flow-based methods \cite{raft, sui2022craft, gaflow} estimate the motion vectors of all pixels between a pair of frames, and then establish point tracks by chaining flow vectors together over multiple frames. This often leads to significant accumulation of error and tracking failure due to occlusions. However, one potential advantage of flow based methods is that they leverage the \emph{spatial continuity} of motion, which arises from the fact that points on the same object often move in a similar way. Recently, CoTracker \cite{cotracker} presented a method to jointly track multiple points and demonstrated that leveraging additional support points in the vicinity of a target point can improve extended point tracking performance. 

\begin{figure*}[t!]
\begin{minipage}[b]{1.0\linewidth}
\centering
\centerline{\includegraphics[width=12.3cm]{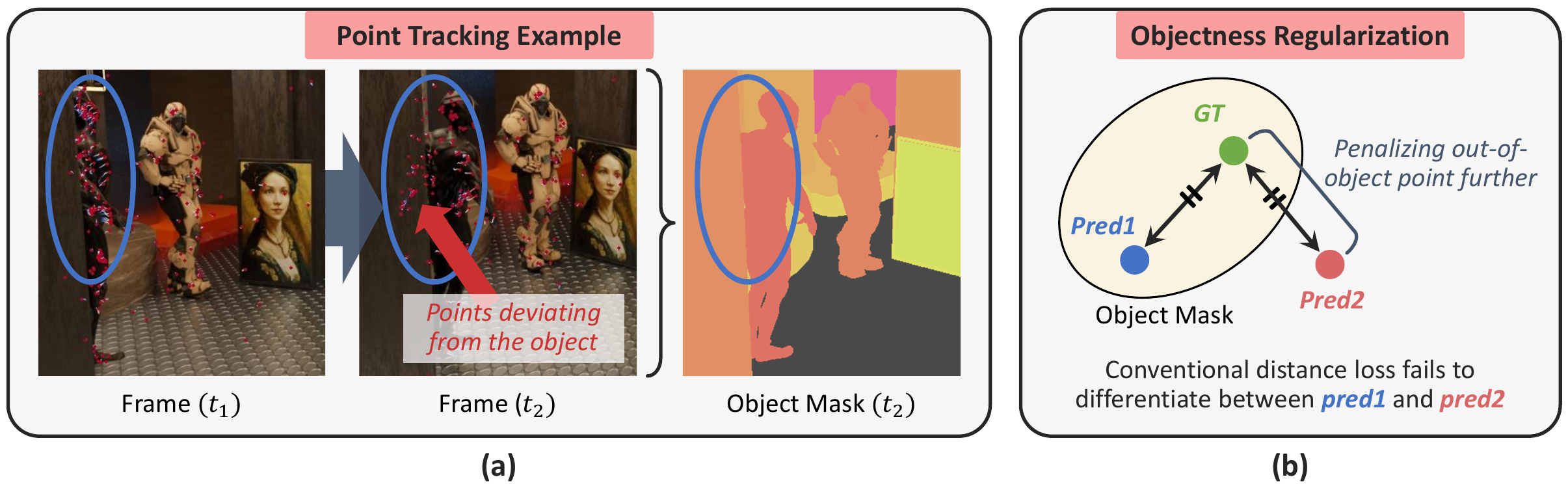}}
\end{minipage}
\caption{(a) shows the example where the points leave the object and fails to return to the location on the original object by missing the object. (b) describes the concept of objectness regularization. Although {\color{red2}{Pred2}} is a worse case than {\color{blue2}{Pred1}} because it misses the object, the conventional distance loss fails to distinguish such cases. To address this, we further penalize the out-of-object points to improve awareness of objectness during the training time.}
\label{figure_teaser}
\end{figure*}

The central thesis of this paper is that the performance of particle video-style point trackers can be significantly improved by leveraging spatial continuity through the inclusion of an \emph{objectness prior}, leading to effective instance-level awareness. Moreover, we show that this can be accomplished by introducing an objectness loss only at training time, which obviates the need for object segmentation at testing time. This allows us to directly incorporate spatial continuity without incurring substantial run-time computational cost, which is extremely beneficial in many applications like AR and robotics. Although prior methods like GAFlow \cite{gaflow} and CoTracker \cite{cotracker} have incorporated  neighborhood information to learn better feature representations, they do not explicitly capture objectness properties in an efficient manner.

The intuition behind our approach is illustrated in Figure \ref{figure_teaser}. Figure \ref{figure_teaser} (a) shows that SOTA trackers frequently predict point correspondences that leave the target object (in this case the figure behind the wall). Once a predicted point deviates from the object it belongs to, it is very difficult for subsequent predictions to return to the correct target object, due to divergence in the modeled appearance. However, access to an object mask makes it clear which object each point belongs to. Our approach to objectness regularization is illustrated in Figure \ref{figure_teaser} (b). For the ground truth point GT, the two predicted locations Pred1 and Pred2 are equally poor matches based on distance (the conventional loss). However, the awareness that Pred1 lies inside the object mask can be captured via an objectness loss, thereby biasing the matcher to prefer points that respect the objectness prior. We show that this significantly improves tracking performance at testing time. By incorporating objectness via a loss at training time, we remove the need to compute object masks at testing time, resulting in a computationally efficient approach. In addition, we leverage contextual attention during point tracking (as was done for optical flow in \cite{gmflow}) to enhance each region feature so that it is aware of neighborhood context. This allows individual objects to be more clearly distinguished from the background or other objects, particularly when they have similar textures or visual patterns. \noindent In summary, the major contributions of this paper are as follows:
\begin{itemize}
	\item We propose an objectness regularization scheme that makes each tracked point aware of the object properties it belongs to. By penalizing predictions that fall outside the object, our approach encourages the points to stay within the object boundaries, leading to effective and efficient long-term tracking.
	\item	We leverage contextual attention to enhance the feature representation for point tracking, enabling each region feature to be aware of its neighborhood context. This enables the model to distinguish individual objects more clearly at the feature level.
	\item	Our approach outperforms existing state-of-the-art methods on three benchmarks: PointOdyssey \cite{pointodyssey}, TAP-Vid-DAVIS \cite{tap}, and CroHD \cite{crohd}. Furthermore, our approach is efficient because the proposed objectness regularization scheme does not require any computational overhead at inference time.
\end{itemize}

\section{Related Works}
\subsection{Optical Flow}
Optical flow aims to precisely estimate the continuous motion of every pixel between two consecutive images, providing a detailed map of movement across the entire scene. Prior works in this domain can be categorized into two main streams:  classical variational approaches~\cite{horn,lucas1981iterative,bruhn2005lucas,black1993framework, elad1998recursive, deriche1995optical, brox2004high} and recently deep learning-based techniques \cite{flownet, xu2017accurate, ilg2017flownet, janai2018unsupervised, raft, flowformer, flowformer++}. Classical approaches, based on assumptions like color constancy and motion smoothness within localized pixel neighborhoods, faced challenges such as the aperture problem and the inability to handle substantial displacements within the scene. In contrast, Teed \emph{et al.} \cite{raft} introduced RAFT, a deep learning paradigm for optical flow estimation. RAFT leverages a 4D correlation volume to compute pixel feature similarity across frames, followed by an iterative update process to estimate the flow. This 4D cost volume approach pioneered by RAFT has been adopted by many subsequent works, not only in optical flow estimation~\cite{xu2022gmflow, jiang2021learning, sui2022craft, lu2023transflow, shi2023videoflow, zhang2021separable} but also in the tracking domain~\cite{pips, tap,tapir, pointodyssey, sun2024refining}. GMFlowNet \cite{gmflow} and GAFlow \cite{gaflow} showed that crafting attention mechanism to capture neighborhood information helps the correspondence matching. {Semantic Optical Flow \cite{sevilla2016optical} leverage the idea that different objects move differently and the optical flow across an image varies depending on the class of the object.}

While optical flow can be used for point tracking by linking estimates across multiple frames, the lack of temporal priors, often restricted to no more than two frames, can lead to significant error accumulation. Although multi-frame optical flow estimation methods \cite{neoral2024mft, shi2023videoflow, chen2023mfcflow} exist, they do not consider points occluded for extended durations, which makes them unsuitable for long-term tracking. In addition, the optical flow work \cite{sevilla2016optical} leveraging object properties requires segmentation masks at inference time and further object classes, which our method does not. Our approach requires object masks only at training time and does not require any class information.
 
\subsection{Point Tracking}
Due to the aforementioned issues of optical flow in video-level tracking, it is required to develop dedicated point tracking methods. In this context, point tracking approaches ~\cite{tomasi1991detection, tap, tapir, 
 pointodyssey, sun2024refining, wang2023tracking, zhou2020tracking} recently attract a lot of attention in both academia and industry fields. {Tomasi \emph{et al.} \cite{tomasi1991detection} developed one of the first methods for point tracking, by matching fixed-sized feature windows in the previous and current frame as the sum of squared intensity differences over the windows.} Doersch \emph{et al.} \cite{tap} introduced TAP-Vid tackling the problem of tracking any point in a video, {followed by TAPIR \cite{tapir} where they show the importance of having good initialization using a matching stage before further refining the estimated point locations in the refinement stage}.  Harley \emph{et al.} \cite{pips} proposed PIPs which utilizes a Particle-Video \cite{particle}-based approach for point tracking, adept at maintaining tracking even through occlusion within a specified temporal window. PIPs++ \cite{pointodyssey} is the improved version of PIPs by adapting to appearance changes of the target via multi-step query features. The aforementioned methods individually track each point; however, the motions of neighboring points are often correlated. Karaev \emph{et al.} \cite{cotracker} introduced CoTracker, a transformer-based one that tracks points jointly by leveraging the correlation between different tracks.

{While we share the use of multi-step query features with PIPs++ and exploit neighborhood information similarly to CoTracker, our approach differs from these methods by considering the object properties to which each point belongs. We propose objectness regularization to improve awareness of objectness by penalizing points outside of their associated objects. Additionally, we incorporate contextual attention to enable the model to effectively distinguish individual objects at the feature level by considering neighborhood contexts.}

\begin{figure*}[t!]
\begin{minipage}[b]{1.0\linewidth}
\centering
\centerline{\includegraphics[width=12.3cm]{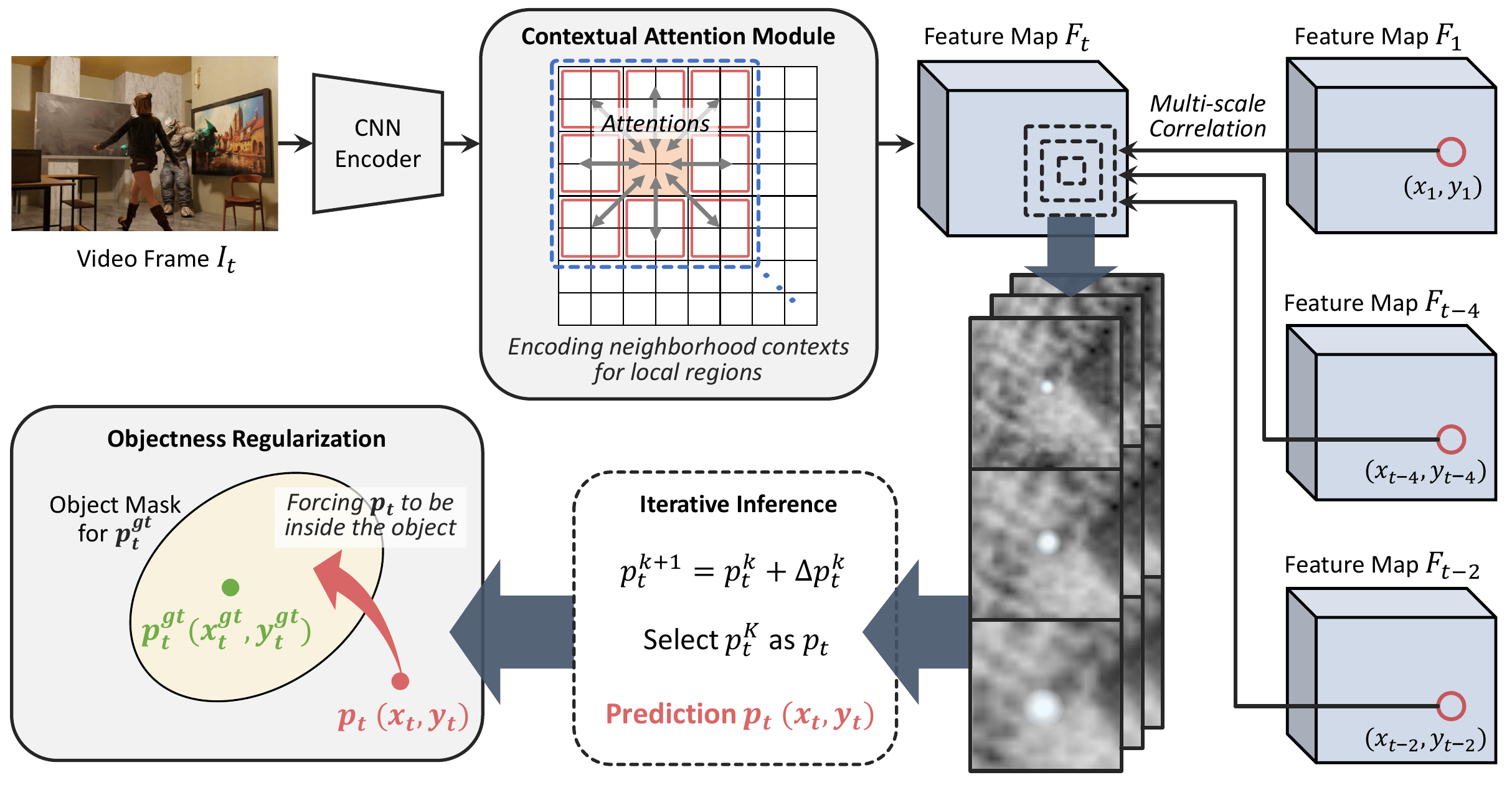}}
\end{minipage}
\caption{Overall framework of our approach at training time. The model consists mainly of feature extraction, iterative inference, and objectness regularization. Contextual attention in the feature extraction improves the representation to better distinguish individual objects by encoding the neighborhood contexts for local regions. The objectness regularization guides the tracked points to stay inside the object by penalizing out-of-object points.}
\label{figure_method}
\end{figure*}

\section{Method}
The point tracking problem can be formulated as follows: Given an input video  $V$ $\in$ $\mathbb{R}^{T \times H \times W \times 3}$ with $T$ frames and an initial point denoted as $p_1$ $\in$ $\mathbb{R}^{2}$ on the first frame, our goal is to predict the corresponding point trajectory $P$ $=$ $\left\{{p}_{t}\right\}_{t=1}^{T}$ $\in$ $\mathbb{R}^{T \times 2}$ throughout the entire video. In this section, we first address the preliminary framework of persistent independent particles (PIPs) \cite{pips,pointodyssey} that our method builds upon. We then present our novel objectness regularization scheme that encourages points to adhere to object boundaries, followed by a contextual attention module that enhances object awareness at the feature level for improved tracking. Figure ~\ref{figure_method} shows the overview of our proposed approach.

\subsection{Preliminary}

The Persistent Independent Particles (PIPs) framework \cite{pips, pointodyssey} tackles the problem of estimating dense point trajectories over a video sequence. The key idea is to track each point independently by leveraging a learned temporal prior and an iterative inference mechanism to refine the trajectory estimates. PIPs++ \cite{pointodyssey}, which our work mainly builds upon, consists of two stages: initialization and iterative updates. 

In the initialization stage, a 2D CNN encoder is used to extract a feature map $F_t$ for each frame $I_t$. The feature representing the appearance of the initial target point $p_1$, denoted as $f_1$, is obtained through bilinear sampling from the first frame feature map $F_1$, at the position corresponding to the point.  All point locations and features in the subsequent $T-1$ frames are then initialized with the first target point location and feature, \{($p_1$, $f_1$)\}.

The iterative update stage aims to refine the trajectory ${P}$ $\in$ $\mathbb{R}^{T \times 2}$ of $p_1$ over $K$ iterations. At each iteration $k$, for each frame $t$, PIPs++ initially extracts local spatial feature crops, around the current estimated point position ${p}_t^k$ from the frame feature $F_t$ at multiple scales. Correlation features between the initial point feature $f_1$ and each feature crop are computed via the dot product. After obtaining correlation features from multiple feature crops, they are concatenated with motion vectors (${p}_t^k - {p}_t^{k-1}$), and then passed through a 1D ResNet to predict position updates $\Delta {p}_t^k$. The new position estimates for the next iteration $k+1$ are then obtained as ${p}_t^{k+1} = {p}_t^{k} + \Delta {p}_t^k$. To adapt to appearance changes, after the first iteration, PIPs++ also correlates feature crops with the recently tracked point features ${f}_{t-2}^k$ and ${f}_{t-4}^k$, in addition to the initial point feature ${f}_1$. Consequently, using the initial point feature $f_1$ preserves the initial appearance of the target point while incorporating $f_{t-2}$ and $f_{t-4}$ contributes to considering recent appearance features, which enables effective tracking despite occlusions and appearance changes.

\subsection{Objectness Regularization}

Each point either belongs to a specific object or the background. Points associated with the same object typically exhibit similar movement patterns and should consistently remain within the object mask. Hence, we propose integrating this underlying object property to mitigate the common issue of points drifting toward different objects or the background while missing the target object. As shown in Figure ~\ref{figure_teaser}, despite both predictions {\color{blue2}{Pred1}} and {\color{red2}{Pred2}} being at an equal distance from the ground truth point location, {\color{blue2}{Pred1}} yields a better prediction due to its placement within the object boundary. Anchoring the predicted points onto the correct object helps avoiding drift towards unrelated objects, which results in more effective long-term tracking.

To this end, we enforce objectness prior in our model via our novel objectness regularization with a training loss $\mathcal{L}_{obj}$, enhancing instance-aware point tracking. We leverage the ground truth object masks from \cite{pointodyssey} for training. In an object mask map, different objects are represented by different values. Specifically, we penalize the model when the predicted point does not belong to the same object mask as the ground-truth point. The loss $\mathcal{L}_{obj}$ for objectness regularization is formulated as:

\begin{equation}
\mathcal{L}_{obj}=\frac{1}{T}\sum_{t=1}^T\mathds{1}\{m^{gt}_{t} \neq m^K_{t}\}\left\lVert p_{t}^{gt}-p^K_{t} \right\lVert_1,
\end{equation}

\noindent where $p_{t}^{gt}$ and $p^K_{t}$ indicate predicted point (at the last $K$ iteration) and ground truth point, respectively. $m^{gt}_{t}$ and $m^K_{t}$ represent the values of the object masks to which $p_{t}^{gt}$ and $p^K_{t}$ belong, respectively. Through the indicator function $\mathds{1}\in\left\{0, 1\right\}$, we can determine whether the predicted point $p^K_{t}$ is inside the target object or not. As a result, we can further penalize the out-of-object points to be inside the corresponding target object by minimizing $\mathcal{L}_{obj}$. In addition to $\mathcal{L}_{obj}$, we employ iterative distance loss function $\mathcal{L}_{dist}$ \cite{pips}, which gives different weights for each iteration step as:

\begin{equation}
\mathcal{L}_{dist}=\sum_{k}^K\gamma^{K-k}(\frac{1}{T}\sum_{t=1}^T\left\lVert p_{t}^{gt}-p^K_{t} \right\lVert_1),
\end{equation}

\noindent where $\gamma < 1$ denotes the weighting term that favors recent update iterations. Since we predict multiple points at the same time, loss functions can be represented as $\mathcal{L}_{dist,i}$ and $\mathcal{L}_{obj,i}$ for $i$-th point among total $N$ tracked points. Combining these two terms, our final training objective can be formulated as:

\begin{equation}
\mathcal{L}=\frac{1}{N}\sum_{i=1}^N(\mathcal{L}_{dist,i}+\alpha\mathcal{L}_{obj,i}), 
\end{equation}

\noindent where $\alpha$ is a hyperparameter for balancing the losses.

\subsection{Contextual Attention}

Feature maps used for matching in motion estimation tasks, such as optical flow and point tracking, need to exhibit two key characteristics. Firstly, they should be locally discriminative. Secondly, they should promote smoothness in motion within the neighborhood, which stems from the observation that nearby points on an object tend to exhibit similar motions. A primary reason for failure in classical optical flow methods \cite{lucas1981iterative} is the reliance on limited context information, which often results in challenges such as the aperture problem~\cite{marr2010vision}. CNN-based feature extractors~\cite{simonyan2014very,he2016deep} employing standard pooling layers to reduce spatial dimension might fail to capture local object boundaries effectively.

To enhance the feature representation for objectness in tandem with the objectness regularization, we leverage a contextual attention module inspired by the optical flow work \cite{gmflow}. The contextual attention encodes neighborhood contexts for local feature regions. As a result, the enhanced feature maps produce sharper peaks in correspondence matching, facilitating the distinction of individual objects even when they have similar visual patterns.

We first extract feature maps from the 2D CNN encoder with $d$ channels and partition them into non-overlapping patches of size $M \times M$ (red squares in the module of Figure \ref{figure_method} indicates patches). Each feature patch then attends to neighborhood $3\times 3$ patches, including itself. For each attention head $j$, we project the vectorized feature patch $Q\in \mathbb{R}^{M^2\times d}$ and the surrounding $3\times3$ vectorized feature patches $V\in \mathbb{R}^{9M^2\times d}$ to dimension $d_{proj}$ using learnable linear projection layers, resulting in $Q_{proj}^j$ and $V_{proj}^j$. We then compute the attention $h_j$, with $Q_{proj}^j$ serving as the query and $V_{proj}^j$ as both the key and value. The outputs of $n$ attention heads $h_j$ are then concatenated and fed through a linear projection layer to produce a feature vector $H\in \mathbb{R}^{M^2\times d}$, where  $d_{proj} = d/n$. These procedures are formulated as follows:

\begin{equation}
h_j = \textit{softmax}(Q_{proj}^j {V_{proj}^j}^{T}/\sqrt{d_{proj}}).V_{proj}^j,
\end{equation}

\begin{equation}
H=\textit{Linear}(\text{concat}([h_1,\dots, h_n])).
\end{equation}

\noindent Then, $H$ replaces the corresponding feature patch region of $Q$. By applying this process in a sliding window manner, we eventually obtain an improved feature map $F_t$ that is aware of the neighborhood context. By incorporating the contextual attention module, our approach enhances the feature representation, making it more effective in distinguishing individual objects based on contextual information.


\section{Experiments}
\subsection{Datasets}
We train our model with the videos of the PointOdyssey training set and evaluate it on the PointOdyssey test set, TAP-Vid-DAVIS, and CroHD datasets following the experimental setting of PIPs++ \cite{pointodyssey}. Following are further details about the point tracking datasets we used:

\subsubsection{PointOdyssey.} PointOdyssey \cite{pointodyssey} dataset is a synthetic benchmark for long-term tracking. This dataset involves around 100 videos with several thousand frames consisting of scenes with both camera and object motion. The test set consists of 12 videos ranging from 884 to 4325 frames in duration.

\subsubsection{TAP-Vid-DAVIS.} TAP-Vid-DAVIS \cite{tap} is a real-world dataset consisting of 30 videos each around 100 frames long with points queried on random objects at random times and during evaluation. {TAP-Vid-DAVIS has uses two evaluation protocols, namely "queried first" and "queried strided". In the "queried first" protocol each point is queried only once which is at the first frame where they become visible, and in the "queried strided" protocol points are queried every five frames with tracking being bidirectional. We evaluate our method on the "query-first" protocol following \cite{pointodyssey}.}

\subsubsection{CroHD.} CroHD \cite{crohd} is a real-world dataset consisting of surveillance-like videos of crowds with tracks annotated on all human heads, with videos varying in length from around 500 frames to a few thousand frames. For evaluation, videos longer than a thousand frames are broken down into thousand frame sequences, yielding a total of 12 sequences.

\begin{table}[t]{
\renewcommand{\arraystretch}{1.25}
\renewcommand{\tabcolsep}{6mm}
\centering
\resizebox{0.85\linewidth}{!}{
\begin{tabular}{l c c c }
\toprule
& \multicolumn{3}{c}{\textbf{PointOdyssey}} \\
\cmidrule(lr){2-4}
\textbf{Method} & $\mathbf{\delta_{avg}}$ $\uparrow$ & \bf Survival $\uparrow$ & \bf MTE $\downarrow$ \\
\bottomrule
RAFT \cite{raft} & 10.1 & 32.6 & 319.5\\
DINO \cite{dino} & 8.6 & 31.3 & 118.4\\
TAP-Net \cite{tap} & 28.4 & 18.3 & 63.5\\
PIPs \cite{pips} & 27.3 & 42.3 & 63.9\\
PIPs++ \cite{pointodyssey} & 29.0 & 47.0& 44.3\\
\cellcolor{cellcol}\bf Ours & \cellcolor{cellcol}\textbf{32.8} & \cellcolor{cellcol}\textbf{52.1} & \cellcolor{cellcol}\textbf{37.6} \\
\toprule
\end{tabular}}
\caption{Performance comparison on the PointOdyssey \cite{pointodyssey} dataset. Our method significantly outperforms all prior methods on all the metrics.}
\label{table_pointodyssey}}
\end{table}

\begin{table}[t]{
\renewcommand{\arraystretch}{1.25}
\renewcommand{\tabcolsep}{6mm}
\centering
\resizebox{0.85\linewidth}{!}{
\begin{tabular}{l c c c }
\toprule
& \multicolumn{3}{c}{\textbf{PointOdyssey}} \\
\cmidrule(lr){2-4}
\textbf{Method} & $\mathbf{\delta_{avg}}$ $\uparrow$ & $\mathbf{\delta^{vis}_{avg}}$ $\uparrow$ & $\mathbf{\delta^{occ}_{avg}}$ $\uparrow$ \\
\bottomrule
PIPs++ \cite{pointodyssey} & 29.0 & 32.4 & 18.8\\
CoTracker \cite{cotracker}&30.2 & 32.7 & \textbf{24.2}\\
\cellcolor{cellcol}\bf Ours & \cellcolor{cellcol}\textbf{32.8} & \cellcolor{cellcol}\textbf{36.3} & \cellcolor{cellcol}23.7 \\
\toprule
\end{tabular}}
\caption{Performance comparison with specific $\delta$ metrics on the PointOdyssey \cite{pointodyssey} dataset. Our method outperforms CoTracker \cite{cotracker} on both $\delta_{avg}$ and $\delta_{avg}^{vis}$. Given the fact CoTracker predicts occlusions while tracking, they slightly outperforms our method on $\delta_{avg}^{occ}$}
\label{PO_cotrack}}
\end{table}

\subsection{Implementation}

\subsubsection{Model Architecture.} We use the same 2D CNN encoder and 8-block 1D Resnet block for position update estimation as PIPs++ \cite{pointodyssey}.The 2D CNN encoder is based on a modified ResNet architecture consisting of  one convolutional layer with 64 kernels followed by 4 layers consisting of 2 residual blocks each, where each layer has 64, 96, 128, 128 kernels respectively. The output from each of these residual layers are concatenated and passed through two more convolutional layers with 256 and 128 kernels respectively, thus producing a feature map with 128 channels and resolution downsampled by a factor of 8. We use the ReLU \cite{nair2010rectified} activation and Instance normalization \cite{ulyanov2017instance} in our encoder. The feature maps from the CNN encoder is further passed through 6 layers of the contextual attention module, with each layer having 8 attention head and using 7 x 7 patches. To compute correlation maps for feature similarity, we compute dot product between the reference features and  feature maps at every timestep at 4 different scales in coarse-to-fine manner. Finally to get the correlation vectors for each point, we sample the correlation maps in a 3 x 3 neighborhood of the estimated point location.
The 1D ResNet module to compute the position updates consists of 1 convolutional layer followed by 8 1D residual blocks and finally a densely connected layer to produce the required position updates for each track.

\subsubsection{Training Details.} We train our model on 140K clips of 24 frames generated from the PointOdyssey \cite{pointodyssey} train dataset. Each clip has a resolution of 384 x 512 and consists of $128$ point tracks. Our model is trained for 300K iterations with a batch-size of $2$ using the AdamW \cite{loshchilov2017decoupled} optimizer and a learning rate of $0.005$ with the 1cycle learning rate policy \cite{smith2019super}. We use the same $\gamma = 0.8$ in $\mathcal{L}_{seq}$ with \cite{pointodyssey} and $\alpha = 0.15$ as weight for our objectness regularlization at training time. Training the model on two RTX 4090 GPUs takes around 2.5 days.

\begin{table}[t!]{
\renewcommand{\arraystretch}{1.25}
\renewcommand{\tabcolsep}{6mm}
\centering
\resizebox{0.85\linewidth}{!}{
\begin{tabular}{l c c c }
\toprule
& \multicolumn{3}{c}{\textbf{TAP-Vid-DAVIS}} \\
\cmidrule(lr){2-4}
\textbf{Method} & $\mathbf{\delta_{avg}}$ $\uparrow$ & \bf Survival $\uparrow$ & \bf MTE $\downarrow$ \\
\bottomrule
RAFT \cite{raft} & 45.2 & 75.4 & 11.5\\
DINO \cite{dino} & 33.1 & 84.1 & 24.6\\
TAP-Net \cite{tap} & 41.73 & 72.92 & 25.93\\
PIPs \cite{pips} & 61.33 & 85.31 & 5.14\\
PIPs++ \cite{pointodyssey}&\textbf{70.5} & 94.0&6.9\\
\cellcolor{cellcol}\bf Ours & \cellcolor{cellcol}69.5 & \cellcolor{cellcol}\textbf{94.6} & \cellcolor{cellcol}\textbf{4.8} \\
\toprule
\end{tabular}}
\caption{Performance comparison on the TAP-Vid-DAVIS \cite{tap} dataset. Our method outperforms prior methods on Survival and MTE metrics showing the effective in long-term tracking on complex real-world scenes.}
\label{table_tap}}
\end{table}

\begin{table}[t]{
\renewcommand{\arraystretch}{1.25}
\renewcommand{\tabcolsep}{6mm}
\centering
\resizebox{0.85\linewidth}{!}{
\begin{tabular}{l c c c }
\toprule
& \multicolumn{3}{c}{\textbf{CroHD}} \\
\cmidrule(lr){2-4}
\textbf{Method} & $\mathbf{\delta_{avg}}$ $\uparrow$ & \bf Survival $\uparrow$ & \bf MTE $\downarrow$ \\
\bottomrule
RAFT \cite{raft} & 15.8 & 62.2 & 82.8\\
DINO \cite{dino} & 8.5 & 37.1 & 116.8\\
TAP-Net \cite{tap} & 22.4 & 35.0 & 60.9\\
PIPs \cite{pips} & 44.0 & 74.9 & \textbf{11.9}\\
PIPs++ \cite{pointodyssey} & 43.4 & 77.5 & 16.4\\
\cellcolor{cellcol}\bf Ours & \cellcolor{cellcol}\textbf{50.9} & \cellcolor{cellcol}\textbf{83.0} & \cellcolor{cellcol}13.5 \\

\toprule
\end{tabular}}
\caption{Performance comparison on the CroHD \cite{crohd} dataset. Our method outperforms prior methods on the $\delta_{avg}$ and Survival metrics.}
\label{table_crohd}}
\end{table}

\subsection{Performance Evaluation}

\subsubsection{Evaluation metrics.} We use the same evaluation metrics used by Zheng et al. \cite{pointodyssey}, namely average position accuracy $\delta_{avg}$, Survival and Median Trajectory Error (MTE). $\delta_{avg}$ was proposed in TAP-Vid \cite{tap} and computed as the average over the percentage of trajectories within a threshold of {1, 2, 4, 8, 16} pixels to the ground truth, in a normalized resolution of 256 x 256. Survival is defined as the ratio of the average number of frames until tracking failure over the video length and failure happens when the L2 distance between the predicted and ground truth trajectory exceeds 50 pixels in the normalized resolution of 256 x 256. The MTE metric measures the median of the distance between the estimated and ground truth tracks.
We evaluate our method with videos at a resolution of 512 x 896 for PointOdyssey dataset \cite{pointodyssey} and Tap-Vid-DAVIS dataset \cite{tap}, and we use a resolution of 768 x 1280 for CroHD dataset \cite{crohd}.

\subsubsection{Compared Methods.} We use point-trackers like PIPS \cite{pips}, TAP-Net \cite{tap}, PIPs++ \cite{pointodyssey}, CoTracker \cite{cotracker}, an optical flow based method RAFT \cite{raft} (where tracks are generated by chaining estimated flows together for consecutive frames) and a feature-matching method DINO \cite{dino} to compare our method against. For RAFT and DINO pretrained weights are used for evalutation while all the other methods are trained with clips from the PointOdyssey training split. We obtain the numbers for the different metrics for PIPS, RAFT, DINO from the PointOdyssey\cite{pointodyssey} paper and for CoTracker from its respective paper. Following CoTracker for fair comparison, the PIPs++ numbers are obtained by using their publicly released official weights and code.


\begin{figure*}[t!]
\begin{minipage}[b]{1.0\linewidth}
\centering
\centerline{\includegraphics[width=12.6cm]{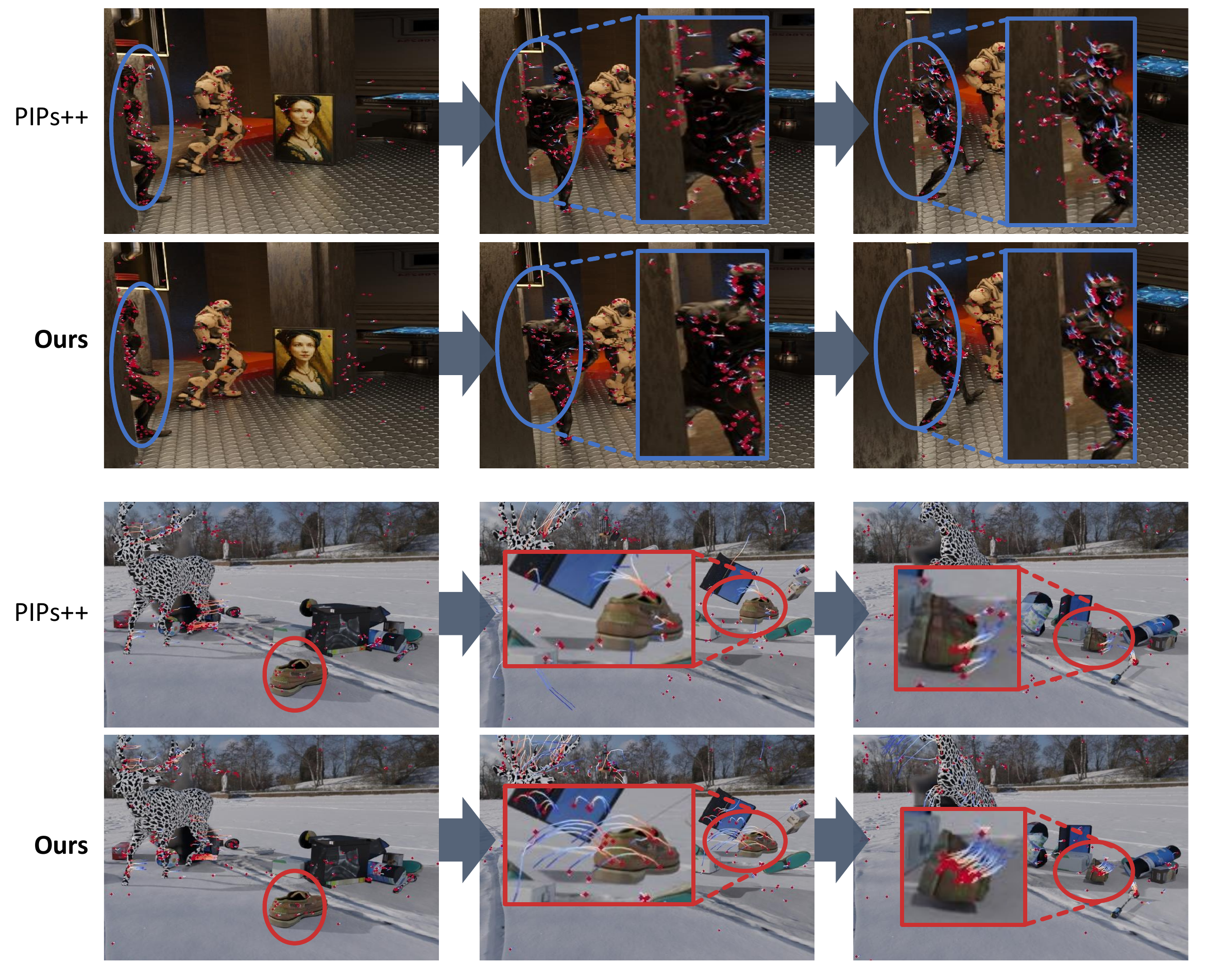}}
\end{minipage}
\caption{Qualitative results demonstrating the benefits of our approach. The examples show cases where our approach tracks the points on each object consistently well.}
\label{figure_qualitative}
\end{figure*}


\begin{figure*}[h!]
\begin{minipage}[b]{1.0\linewidth}
\centering
\centerline{\includegraphics[width=12.3cm]{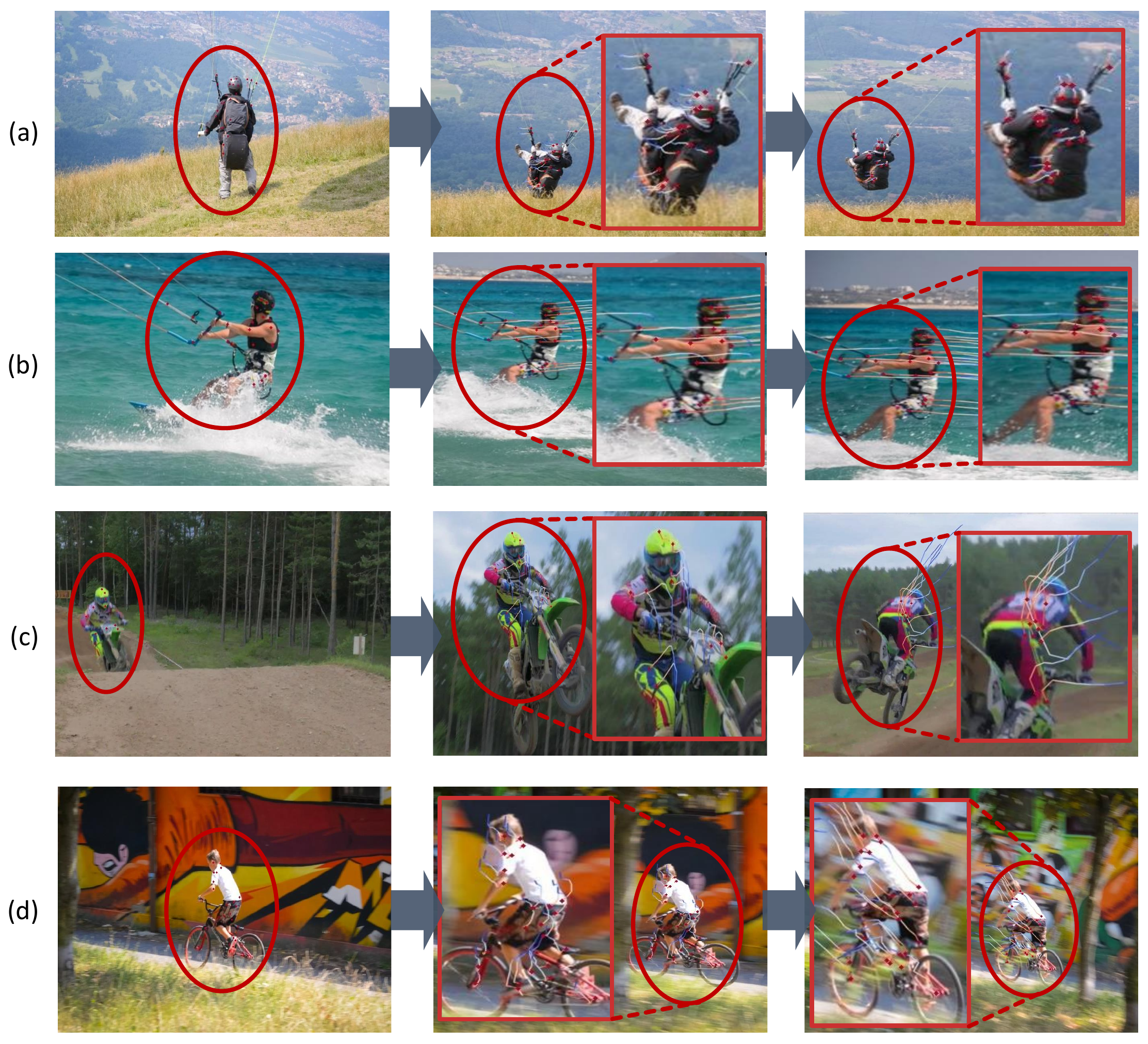}}
\end{minipage}
\caption{Qualitative results on the TAP-Vid-DAVIS \cite{tap} dataset. Ours can effectively track points in various scenarios with occlusion, motion blur and changing orientations}
\label{figure_tap}
\end{figure*}


\subsubsection{Performance comparisons.} Table \ref{table_pointodyssey}, \ref{table_tap} and \ref{table_crohd} shows tracking performance comparison with $\delta_{avg}$,  Survival, and MTE metrics on the PointOdyssey, TAP-Vid-DAVIS, and CroHD datasets, respectively. Our method overall outperforms existing prior methods, showing the effectiveness of our proposed designs. Table \ref{PO_cotrack} shows the comparison results with specific $\delta$ metrics (\emph{i.e.}, $\delta_{avg}$,  $\delta_{avg}^{vis}$,  $\delta_{avg}^{occ}$), where the later two are similar to $\delta_{avg}$, but with only visible or occluded points. We outperform both methods on the first two metrics and achieve competitive performance for $\delta_{avg}^{occ}$.

\subsection{Qualitative Results}

Figure \ref{figure_qualitative} shows qualitative results with visualized points from PIPs++ and ours. As shown in the first two rows of images, the animated humanoid (circled in blue) becomes occluded by a wall of similar color in the initial frame, making it a very challenging scenario. In the following frames, predicted points from PIPs++ fails to stay within the target humanoid but drift away to the wall. In contrast, our approach prevents such drift and keeps tracking the underlying object they belong to. For the below example, our approach tracks the points on the shoe (circled in red) well while PIPs++ fails to do so. In the case of PIPs++, many points leave the object (\textit{i.e.}, shoes) and fail to return to the correct target object.

Figure \ref{figure_tap} shows visualization results on the TAP-Vid-DAVIS \cite{tap} dataset which includes real-world video samples. Our method can effectively track the points on the paraglider (a) even under sudden change of orientation of the person and can even track the single point on the very thin rope. (b) and (c) are demostrations of effective tracking by our method under fast motion and also under complete change of viewpoint in the case of the biker. In (d) we can see our model can track points consistently even under motion blur and over come occlusions as the cyclist crosses the tree-barks in the view. These examples show the effectiveness of our method in tracking points in real-world videos with diverse motion.

\subsection{Effect of Proposed Designs}
Table \ref{table_ablation} shows the effect of our "Objectness Regularization" and "Contextual Attention" with the baseline model, PIPs++. As shown in the table, both objectness regularization and contextual attention components properly contributes to the point tracking performances (\textit{i.e.}, $\delta_{avg}$, Survival, MTE). As a result, we achieve the best performances on all the evaluation metrics with our final model including both objectness regularization and contextual attention. Note that the objectness regularization does not require any computational overhead. It is only applied at training time. In addition, the contextual attention module only requires a small number of network parameters. As a result, the parameter number of our baseline is 17.6M while the parameter number of our proposed method is 18.6M. This gap is quite marginal, but the performances are significantly improved by the proposed designs.

\begin{table}[t]{
\renewcommand{\arraystretch}{1.3}
\renewcommand{\tabcolsep}{4mm}
\centering
\resizebox{0.99\linewidth}{!}{
\begin{tabular}{c c c c c }
\toprule
\multicolumn{2}{c}{\textbf{Proposed Designs}} & \multirow{2}{*}{$\mathbf{\delta_{avg}}$ $\uparrow$} & \bf \multirow{2}{*}{Survival $\uparrow$} & \bf \multirow{2}{*}{MTE $\downarrow$} \\
\bf \makecell{Objectness Reg.} & \bf \makecell{Contextual Att.} &  &  &  \\
\bottomrule
\xmark & \xmark  & 29.0 & 47.0 & 44.3\\
\cmark   & \xmark  & 30.2 & 49.6 & 42.3\\ 
\xmark  & \cmark  & 30.6& 49.1 &48.3\\ 
\cellcolor{cellcol}\cmark & \cellcolor{cellcol}\cmark & \cellcolor{cellcol}\textbf{32.8} & \cellcolor{cellcol}\textbf{52.1} & \cellcolor{cellcol}\textbf{37.6}\\
\toprule
\end{tabular}}
\caption{Effects of the proposed designs on the performances for PointOdyssey. Both designs fairly improve tracking performance over baseline.}
\label{table_ablation}}
\end{table}

\begin{table}[t]{
\renewcommand{\arraystretch}{1.3}
\renewcommand{\tabcolsep}{5.5mm}
\centering
\resizebox{0.9\linewidth}{!}{
\begin{tabular}{c c c c c  c}
\bottomrule
${\alpha}$ & 0 & 0.05 &  \cellcolor{cellcol}0.15 \cmark &  0.5 & 1\\ \midrule
 \textbf{Survival} $\uparrow$ & 49.1 & 50.3 & \cellcolor{cellcol}\textbf{52.1} & 50.6& 49.8 \\	\toprule
\end{tabular}}
\caption{Survival performances on the PointOdyssey dataset according to different weight $\alpha$ for our objectness regularization.}
\label{table_hyperparameter}}
\end{table}

\subsection{Effect of Regularization Weight}

 Table \ref{table_hyperparameter} shows Survival performances based on the weight $\alpha$ for our objectness regularization (please refer to equation (3) in Section 3.2). Note that $\alpha$ adjust the weight of objectness regularization compared to the typical distance loss. As shown in the table, we could obtain higher performance than the existing methods with any weight $\alpha$ values. In particular, we achieve the best result when using $\alpha=0.15$.

\section{Discussion}

Our work demonstrates the effectiveness of learning object priors in point tracking by utilizing object masks which are readily available in synthetic environments. Synthetic data is mostly used in practice for training point tracking methods due to the ease of obtaining point correspondence labels. However, there can be a domain gap between synthetic data and real-world data, thus training with further real-world data could be beneficial for practical real-world applications. Future work could explore the use of object masks generated by foundation models like Segment Anything \cite{kirillov2023segment} to extend our method to real-world data training, potentially bridging this domain gap.

\section{Conclusion}
In this work, we introduce a novel object-aware approach for point tracking that encourages tracked points to stay within the boundaries of object instances. Our key ideas include an objectness regularization scheme that penalizes points drifting outside their associated objects during training, and a contextual attention module that enhances feature representations to better distinguish individual objects. Extensive experiments on the PointOdyssey, TAP-Vid-DAVIS, and CroHD benchmarks demonstrate the effectiveness of our approach, with state-of-the-art performance across multiple evaluation metrics. Ablation studies confirm the complementary benefits of the objectness regularization and contextual attention components. Our approach substantially improves tracking robustness and accuracy without sacrificing efficiency.

\noindent{\bf Acknowledgement.} Portions of this work were supported by a grant from Toyota Research Institute under the University 2.0 program.

\bibliographystyle{splncs04}
\bibliography{main}
\end{document}